%% file: ccl2024-en.tex
\documentclass[11pt]{article}
\usepackage[hyperref]{ccl2024-en}
\usepackage{times}
\usepackage{url}
\usepackage{latexsym}
\usepackage{fancyhdr}

\usepackage{graphicx}   
\usepackage{CJKutf8}  

\usepackage{pgfplots}   
\usepgfplotslibrary{groupplots}  
\usetikzlibrary{patterns}   
\usepackage{caption,subcaption}  

\usepackage{indentfirst} 
\usepackage{booktabs}  
\usepackage{multirow}  
\usepackage{amsmath}   
\usepackage{caption} 
\usepackage{wrapfig} 
\usepackage{makecell} 
\usepackage{xspace}

\newcounter{mycount}        

\usepackage{colortbl}
\definecolor{myblue}{RGB}{169, 232, 251}
\definecolor{myyellow}{RGB}{255, 217, 178}

\definecolor{case_red}{RGB}{207, 16, 32}  
\definecolor{case_blue}{RGB}{0, 84, 159}  
\definecolor{prompt_blue}{RGB}{168, 217, 224}   

\title{Translate-and-Revise: Boosting Large Language Models \\ for Constrained Translation}

\author{Pengcheng Huang\textsuperscript{1}\thanks{\xspace\xspace Equal contribution.}, Yongyu Mu\textsuperscript{1}\footnotemark[1], Yuzhang Wu\textsuperscript{1}, Bei Li\textsuperscript{1}, \\
{\bf  Chunyang Xiao\textsuperscript{3},Tong Xiao\textsuperscript{1,2}\thanks{\xspace\xspace Corresponding author.}, \and Jingbo Zhu\textsuperscript{1,2}} \\
	\textsuperscript{1}NLP Lab, School of Computer Science and Engineering, \\ Northeastern University, Shenyang, China\\
	\textsuperscript{2}NiuTrans Research, Shenyang, China\\
    \textsuperscript{3}JP Morgan, United Kingdom\\
    \ttfamily{hpc1449181552@outlook.com lixiaoyumu9@gmail.com}\\
	\ttfamily{\{xiaotong,zhujingbo\}@mail.neu.edu.cn} 
}

\date{}

\begin{document}
\maketitle
\vspace{1em}
\begin{abstract}
  Imposing constraints on machine translation systems presents a challenging issue because these systems are not trained to make use of constraints in generating adequate, fluent translations. In this paper, we leverage the capabilities of large language models (LLMs) for constrained translation, given that LLMs can easily adapt to this task by taking translation instructions and constraints as prompts. However, LLMs cannot always guarantee the adequacy of translation, and, in some cases, ignore the given constraints. This is in part because LLMs might be overly confident in their predictions, overriding the influence of the constraints. To overcome this overiding behaviour, we propose to add a revision process that encourages LLMs to correct the outputs by prompting them about the constraints that have not yet been met. We evaluate our approach on four constrained translation tasks, encompassing both lexical and structural constraints in multiple constraint domains. Experiments show 15\% improvement in constraint-based translation accuracy over standard LLMs and the approach also significantly outperforms neural machine translation (NMT) state-of-the-art methods.
\end{abstract}

\section{Introduction}
\label{intro}

%
%

\cclfootnote{
    %
    %
    \hspace{-0.65cm}  
    \textcopyright 2024 China National Conference on Computational Linguistics

    \noindent Published under Creative Commons Attribution 4.0 International License
}

Constrained translation seeks to generate translations that adhere to pre-specified constraints. To achieve this, conventional approaches impose constraints on machine translation systems and force them to follow the constraints during inference~\cite{hokamp2017lexically,hasler2018neural,dinu2019training,bergmanis2021facilitating,wang2022integrating,ailem2021encouraging}. More recently, large language models (LLMs) have been shown to be strong translation systems~\cite{hendy2023good,DBLP:conf/eamt/MoslemHKW23}. They provide a general way to involve various instructions, demonstrations, and constraints into the translation process~\cite{mu2023augmenting,bogoychev2023terminology}, enabling us to perform constrained translation using off-the-shelf, well-trained LLMs.

\begin{wrapfigure}{r}{0.5\textwidth}
  \centering
  \vspace{-2em}
  \includegraphics[width=0.5\textwidth]{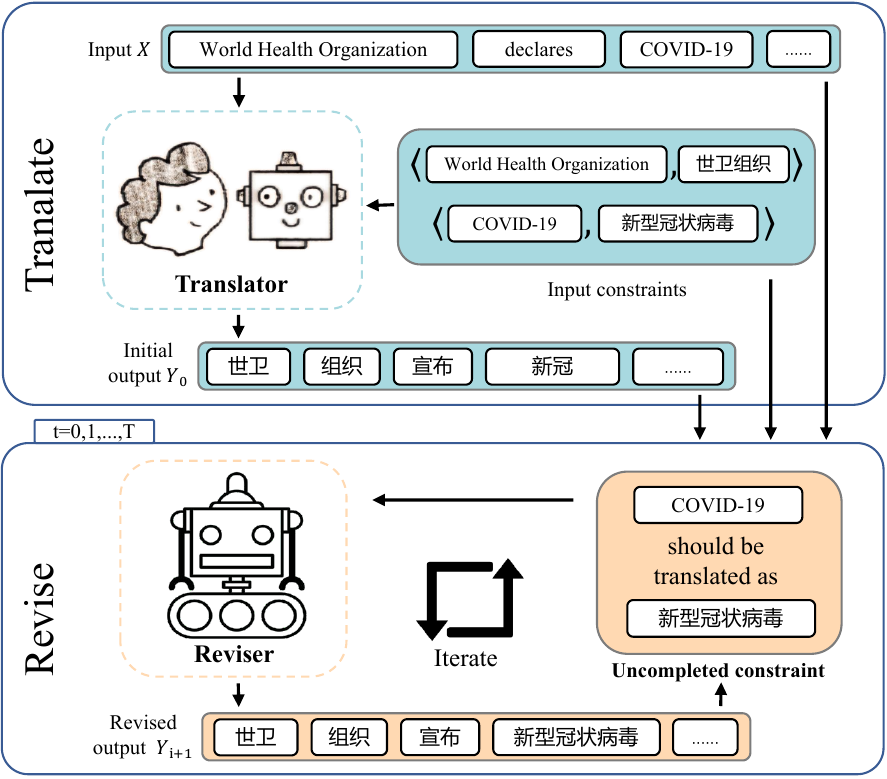}
  \caption{
Given source language input $X$ and constraint pairs, a \textit{Translator} produces an initial translation $Y_0$ where COVID-19 is translated as ``\begin{CJK*}{UTF8}{gbsn}新冠\end{CJK*}''. Subsequently, a \textit{Reviser} iteratively revise the translation $Y_i$ to a better one $Y_{i+1}$, correctly translating COVID-19 as ``\begin{CJK*}{UTF8}{gbsn}新型冠状病毒\end{CJK*}''.
  }
  \label{fig:f1}
  \vspace{-1.9em}
\end{wrapfigure}

While applying LLMs to constrained translation is straightforward, we observe empirically that even strong LLMs (i.e. \texttt{GPT-3.5}) do not always follow the instructions to obey constraints: 
LLMs' predictions often override the guide of constraints, which result in missing constraints during translation. See Figure \ref{fig:f1} for an example where we use an LLM to translate an English sentence to a Chinese sentence with a lexical constraint ``COVID-19$\to$\begin{CJK*}{UTF8}{gbsn}新型冠状病毒\end{CJK*}''. We note that, despite significant effort in developing clear and instructive prompts, we were not able to improve the LLM in a single run of the LLM through the use of these constraints. For instance, we observed that when using open-source LLM to translate COVID-19, it tends to translate it as ``\begin{CJK*}{UTF8}{gbsn}新冠\end{CJK*}'' more than 80\% of the time, overlooking the constraint in the prompt to translate COVID-19 as ``\begin{CJK*}{UTF8}{gbsn}新型冠状病毒\end{CJK*}''. The problem consists of a real use case for what describes as `memo trap' in the LLM literature~\cite{mckenzie2023inverse}.

To alleviate this problem and thus improve the accuracy to meet constraints, we propose to construct prompts iteratively that enable better focus on the unsatisfied constraints. The idea behind our approach is to leverage the auto correction skills of LLMs by explicitly prompting them with which constraints are not satisfied~\cite{DBLP:journals/corr/abs-2303-17651,DBLP:conf/acl/ZhangLLLJ23,DBLP:journals/corr/abs-2306-02907}. To do this, we introduce a revision step after the initial run of LLMs where we provide the LLMs with both the already-generated translation and the constraints that have not been covered. Then, we instruct the LLM to revise its output by taking these constraints into account. 

We conduct experiments across four diverse constrained translation datasets, encompassing two distinct constraint types: lexical and structural. Our proposed ``Translate-and-Revise'' (TAR) approach consistently elevates the performance of LLMs in constrained translation, achieving state-of-the-art (SoTA) results on multiple datasets.

The contributions of this work are as follows:
\begin{itemize}
\item We introduce a novel TAR strategy that initially employs LLMs as constraint-aware translators and subsequently reproposes them as revisers to revise translations that do not meet given constraints. We show that TAR significantly reduces missing constraints during translations. 
\item We rigorously evaluate our approach on four constrained translation datasets spanning multiple domains like news and electronics. Our results demonstrate a significant improvement in constraint fidelity and translation quality, outperforming existing methods and achieving SoTA results.
\item To the best of our knowledge, our study is the first to evaluate LLMs across four distinct constrained translation datasets, thereby providing a robust LLM baseline for future research in the area. We believe our findings serve a solid baseline towards establishing more comprehensive benchmarks in the field of constrained translation.
\end{itemize}

\section{Methods}
Given a source language input and bilingual constraints, TAR first employs LLMs as translators for an initial translation.  While this step often yields high-quality outputs, when the LLMs' confidence during generation exceeds the guidance of the constraints, it results in suboptimal translation outputs. To mitigate the occurrence of missing constraints in LLMs-based translation, we introduce a reviser to enhance adherence to the constraints in the translation. The revision process is iterated multiple times until all constraints are satisfied, or the maximum allowable number of modifications is reached. The process of TAR is provided in Figure \ref{fig:f1}. Next, we describe TAR in more details.

\subsection{Translate}
Let $X = \{x_1, x_2, ..., x_n\}$ be the source-language sentence with length $n$, and $Y = \{y_1, y_2, ..., y_m\}$ be the target-language sentence with length $m$. The translation procedure can be written as:

\begin{equation}
Y = \mathrm{Trans}(f(X))
\label{equation_1}
\end{equation}
where $\mathrm{Trans}(\cdot)$ symbolizes the translation model (either an NMT model or an LLM), and $f(\cdot)$ denotes a template by which we process $X$ to make it suitable as the input of $\mathrm{Trans}(\cdot)$. 

Let the $\langle S, T\rangle = \{\langle s_1,t_1\rangle, \langle s_2,t_2\rangle, ..., \langle s_k,t_k\rangle\}$ represents the bilingual constraints with $k$ pairs in total. Constrained translation needs the system to accurately translate each source constraint $s_i$ to its corresponding target constraint $t_i$. This process can be represented by the following equation:

\begin{figure*}
\centering
\input{tables/prompt.tex}
\caption{Two stages of TAR, in the Translate stage, constraints $\langle S,T \rangle$ are incorporated into the prompt to enable the model to generate preliminary translation results that meet the constraints to a certain extent. In the Revise stage, LLMs revise the flawed translation results $Y^{flawed}$ with uncompleted constraints $\langle S,T \rangle^{un}$. The sections shaded in blue \colorbox{prompt_blue}{\rule{0cm}{0.25cm}} and yellow \colorbox{myyellow}{\rule{0cm}{0.25cm}} respectively represent the important parts of the two stages.}
\label{fig:f2}
\vspace{-2.5em}
\end{figure*}
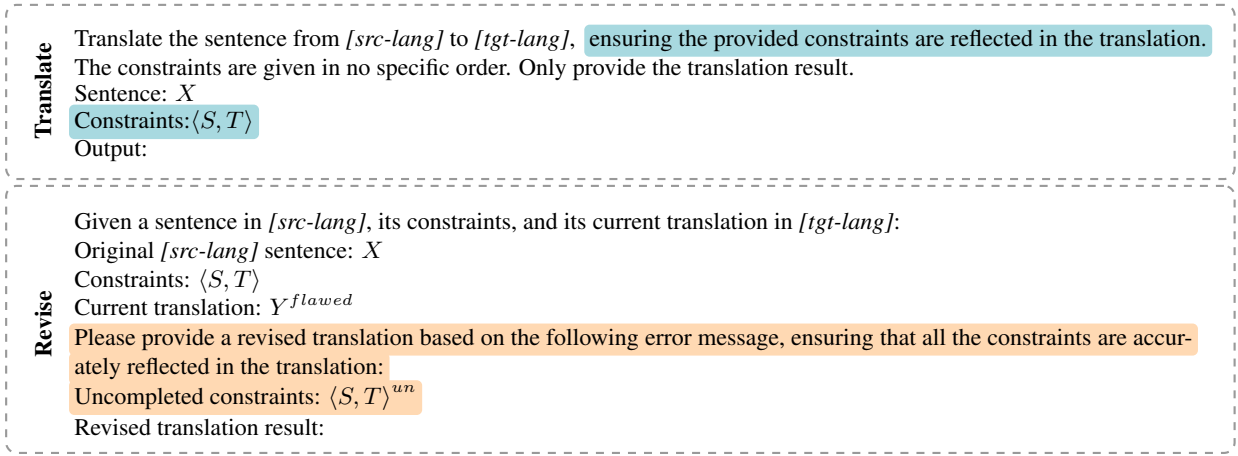

\begin{equation}
\begin{split}
Y = \mathrm{Trans}(&f(X, \langle S, T\rangle)) \\
s.t.\enspace T &\in Y
\end{split}
\label{equation_2} 
\end{equation}

Since conventional translation instructions never impose constraints on LLMs, they frequently fall short of satisfying constraints. In this work, we propose to integrate these constraints directly into the prompts and employ an instruction based on natural language specifically tailored for constrained translation tasks. Our template $f(\cdot)$ is shown in Figure \ref{fig:f2} which can effectively turn LLMs into constraint-aware translators.

\subsection{Revise}
However, the LLM-based translation cannot always cover all original constraints. We randomly sampled 20 incorrect translation results and observed that, in datasets like WMT21 Terminology Translation \cite{DBLP:conf/wmt/AlamKABDFGJKN21}, to $95\% (19/20)$ of the cases, the tokens generated by the model were similar to the expected constraints meaning and exhibited high confidence levels. The confidence level of LLMs in generating these tokens remained virtually unchanged, whether or not constraints were included in the instructions, revealing overconfidence in generation while overlooking the constraints.

We notice a strong connection between our real use case and `memo trap'~\cite{mckenzie2023inverse} as unsatisfied constraints often pertain to non-mainstream translations resulting terms used with lower frequency and the incorrect translations usually refer to the mainstream translations. Compared `memo trap', we show that the phenomenon extends to toy settings and is prominent even for real applications and for SOTA models like GPT-3.5. 


To overcome these challenges, we initially employ a rule-based method to identify which constraints are not completed. Subsequently, these uncompleted constraints ${\langle S,T \rangle}^{un}$, along with the source language input $X$, flawed translations output $Y^{flawed}$, and all other given constraints $\langle S, T\rangle$, are passed to the LLM. At this juncture, the LLM assumes the role of a reviewer, tasked with revising flawed translation upon receipt of uncompleted constraints. The aforementioned process is defined by the following formula:

\begin{equation}
Y = \mathrm{Revise}(f(X, \langle S,T \rangle , {\langle S,T \rangle}^{un}, Y^{flawed}))
\label{equation_3}
\end{equation}
where $\mathrm{Revise}(\cdot)$ symbolizes the reviser. Furthermore, TAR can continuously iterate in the loop of detecting uncompleted constraints and making revisions until a stopping condition is met. This condition is either the iteration $i$ reaches a specified count or the translation satisfies all constraints. We represent this iterative process as follows:

\begin{equation}
Y_{i+1} = \mathrm{Revise}(f(X, \langle S,T \rangle , {\langle S,T \rangle}^{un}_{i}, Y_{i}))
\label{equation_4}
\end{equation}

where the translation $Y_{i}$ from the previous iteration, combined with its uncompleted constraints ${\langle S,T \rangle}^{un}_{i}$ and other inputs, is sent into the reviser to produce a more precise translation $Y_{i+1}$. By highlighting uncompleted translation constraints and comparing flawed translation results, human translators are able to satisfy these constraints and optimize translation output. Empirically, we find that LLMs can revise the translation results similarly to human translators, while being more efficient and cost-effective. Additionally, we discuss the time and financial costs of multiple iterations in Appendix \ref{appendix_a}.

\section{Experiments}
In this study, we evaluate the performance of the TAR in constrained translation. While most previous research has typically focused on just one or two constrained translation tasks~\cite{dinu2019training,wang2022template,hashimoto2020high,zeng2023extract}, our evaluation expands to  two types of constraints: lexical constraints and structural constraints, covering four practical scenarios: general lexically constrained translation, translation with terminology constraints,  translation with named entity constraints, and structured document translation. 

\input{tables/data_description}

\subsection{Setup}
\textbf{Datasets} 
Detailed information about the datasets we used can be found in Table \ref{tab:data_description}. Lexical constraints refer to sentences with predefined word or phrase constraints sourced from existing databases. Structural constraints, in contrast, encompass inline markup tag constraints like XML tags, for example, \texttt{\textless ph\textgreater} and \texttt{\textless/ph\textgreater}. Here are more details about the datasets that we use in this work:

\textit{General Lexically Constrained Translation:} This is based on a dataset\footnote{\url{https://github.com/mtresearcher/terminology_dataset}} provided by~\cite{dinu2019training}. The dataset is derived from \textit{newstest2017} En $\to$ De. Lexical constraints are extracted with guidance from two general-domain term databases: IATE and Wiktionary\footnote{Available at \url{https://iate.europa.eu/home} and \url{https://www.wiktionary.org/}.}.

\textit{Terminology Translation:} To benchmark against SoTA NMT systems, we employ the official test set from the \textit{terminology translation} task in WMT21\footnote{\url{https://www.statmt.org/wmt21/terminology-task.html}}(WMT21 TT)~\cite{DBLP:conf/wmt/AlamKABDFGJKN21}.

\textit{Entity Translation:} We also endeavor to evaluate our method using the extensive Entity Translation Corpus (ETC)~\cite{zeng2023extract}, which comprises six test sets from the WMT News Translation Task spanning 2015-2021. For alignment, we employ spaCy NER models\footnote{\url{https://pypi.org/project/spacy/}} to extract source entities and use awesome-align~\cite{dou2021word} for their correspondence.

\textit{Structured Document Translation:} Following recent works~\cite{wang2022template}, we conduct experiments on the LXM dataset\footnote{\url{https://github.com/salesforce/localization-xml-mt}}~\cite{hashimoto2020high}, in which XML tags are hierarchically distributed throughout the source and target text.

\textbf{Evaluation Metrics}
Consistent with previous studies~\cite{dinu2019training,ailem2021encouraging,wang2022template,zeng2023extract}, we employ BLEU~\cite{papineni2002bleu} and the constraints completion rate (CCR) to assess translation quality and constraint-based translation accuracy except for structured document translation. For entity translation, we also incorporate the COMET score\footnote{\texttt{wmt22-comet-da}}~\cite{rei2020comet} for comparison with the work of Zeng~\shortcite{zeng2023extract}. In structured document translation, we utilize sacreBLEU~\cite{post2018call} to compute the XML-based BLEU score\footnote{XML tags are treated as an integral part of the sentences during BLEU score calculation.}. Additionally, we measure structured constraint-based translation accuracy using the structure accuracy rate (SAR) and structure match rate (SMR). Here, SAR evaluates the compatibility of translation results with XML parsers, while SMR ensures the translated XML structure aligns with the reference. Both metrics are assessed using lxml\footnote{\href{https://lxml.de/}{https://lxml.de/}}.

\input{tables/f1_and_f2}

\textbf{Baselines}
To ensure thorough evaluation, apart from comparing TAR with the LLM baseline without revision, we also we compare TAR with representative methods across each of the four tasks. For general lexically constrained translation, our baselines include the vanilla Transformer~\cite{vaswani2017attention}, Const.Dec.~\cite{post2018fast}, Code-switching~\cite{dinu2019training}, Append~\cite{dinu2019training}, and Robust Terminology Translation (RTT)~\cite{zhang2023understanding}.
For the terminology translation task, our baselines are derived from the top three submissions of WMT21. They include HW-TSC~\cite{wang2021huawei}, Term-Mind-sys2~\cite{wang2021termmind}, ProMT.soft~\cite{molchanov2021promt}, TildeMT~\cite{bergmanis2021dynamic}, and Lingua Custodia~\cite{ailem2021lingua}.
For entity translation, we utilize the vanilla Transformer, Code-switching, Placeholder~\cite{yan2019impact}, and Extract and Attend~\cite{zeng2023extract}.
Structured document translation baselines consist of the vanilla Transformer, Split-Inject~\cite{al1997automatic}, and Template~\cite{wang2022template} methods.

\textbf{Model Configurations}  
We initially investigate the potential of LLMs to act as both the translator and reviser within TAR. Our primary choice for LLMs is \texttt{gpt-3.5-turbo-0613}\footnote{\url{https://openai.com/}}, chosen for its exceptional translation capabilities and proficiency in adhering to instructions. Additionally, we assess its revising process for NMT models in Section \ref{TAR_nmt} and explore whether TAR provides consistent improvements across different LLMs in Section \ref{TAR_LLMs}. The decoding parameters for these models remain at their default settings, except for the sampling temperature, which is set to 0. We employ natural language-based prompts in a one-shot manner, merging uncompleted constraints with the source language, flawed translation results, and original constraints to form the reviser's input. These prompts are depicted in Figure \ref{fig:f2}.

\input{tables/entity_res}

\input{tables/struct_res}

\textbf{Detection of Uncompleted Constraints}
To identify unmet constraints in translations, we employ a rule-based procedure that leverages scripts designed for calculating CCR. This procedure assesses how well the translation adheres to the constraints. We further explore the capacity of LLMs to autonomously verify constraint completion and offer detailed feedback to the reviser in Section \ref{sec:reviser_input}.

\subsection{Main results}
Table \ref{tab:dinu_res}, Table \ref{tab:term_res}, Table \ref{tab:entity_res}, and Table \ref{tab:structure_res} detail the performance of TAR on general lexically constrained translation, terminology translation, entity translation, and structured document translation, respectively. Here are our main results.

\textbf{Comparison with base LLMs}
TAR consistently boosts the performance of LLMs in constrained translation. Two primary factors contribute to this improvement:

(1) Our natural language-based prompts, as opposed to the conventional few-shot translation prompts~\cite{hendy2023good}, are more effective for constrained translation. Specifically, in terminology translation (refer to Table \ref{tab:term_res}), our prompts lead to an average BLEU score increase of 0.3 and a CCR rise of 7.7\%.

We noticed significant gains over base LLMs in entity translation. Across four language directions, there is a consistent uplift in BLEU scores, averaging an increase of 0.9. Notably, the CCR experiences increases of 19.4\%, 8.9\%, 24.3\%, and 19.9\% respectively. This marked improvement can primarily be attributed to the superior instruction-following capabilities of LLMs. By incorporating constraints in instructions, we can guide the model more effectively to address these constraints, alleviating the issue of LLMs struggling to correctly translate named entities.

(2) Revision effectively improves constraint-based translation accuracy across all datasets without sacrificing translation quality, addressing the tendency of LLM-based translations to overlook constraints. 

\input{tables/case_study}

As illustrated in Table \ref{tab:case_study}, we observe that the LLM exhibits overconfidence in its prediction overriding the influence of constraints, especially when the constraint suggests an uncommon translation for a polysemous word. For instance, the LLM translates ``COVID-19'' to the more commonly used ``\begin{CJK*}{UTF8}{gbsn}新冠\end{CJK*}'' instead of adhering to the target constraint ``\begin{CJK*}{UTF8}{gbsn}新型冠状病毒\end{CJK*}''. We speculate this overconfidence in LLMs stems from their greater exposure to ``\begin{CJK*}{UTF8}{gbsn}新冠\end{CJK*}'' compared to ``\begin{CJK*}{UTF8}{gbsn}新型冠状病毒\end{CJK*}'' during the pre-training phase, which may lead them to be overly loyal to certain patterns, thereby preventing them from meeting certain constraints.

However, the revision step can effectively cut down the possibility of missing constraints by explicitly prompting LLMs with which constraints are not satisfied. Experimental results indicate that our revision strategy led to average improvements in CCR by 2.2\% for lexically constrained translation, 2.5\% for terminology translation, and 1.5\% for entity translation across various language directions. While the improvements in SCR and SMR for structured document translation might not seem prominent, it's primarily because the initial translation is already at a 100\% performance.

\textbf{Comparison with supervised methods}
The SoTA methods for these four constrained translation datasets predominantly rely on pseudo-data augmentation. Through our experimental results, we observe that these methods nearly achieve perfection on the IATE, Wiktionary, and LXM datasets (as evidenced in Table \ref{tab:dinu_res} and Table \ref{tab:structure_res}). We contend that the test sets for these datasets might be relatively straightforward, and the constraints they encompass are frequently encountered in training sets. Therefore, they may not accurately reflect real-world applications where constraints might span multiple domains and are infrequently seen in the training data. However, when assessed on the ETC (as depicted in Table \ref{tab:entity_res}), which comprises test data spanning from 2015 to 2021, showcasing a rich diversity in constraint domains, the efficacy of traditional data augmentation methods seems to be poor. In contrast, TAR's performance remains stable, demonstrating comparable constraint-based translation accuracy on ETC as with other datasets. This highlights TAR's proficiency in handling the diverse constraint requirements found in real-world situations. 

\subsection{Impact of Inputs on Reviser Performance}\label{sec:reviser_input}

\input{tables/feedback_ablation}

The reviser receives inputs including the source language sentence, translation results, given constraints, and the uncompleted constraints. To evaluate the significance of each component, we conducted experiments wherein we omitted specific elements from the input. The variations include: 1) Excluding uncompleted constraints; 2) Excluding original constraints; and 3) Only indicating to the model that the translation is flawed without specifying the uncompleted constraints. All other settings remain unchanged. The comparative outcomes on IATE and Wiktionary are presented in Table~\ref{tab:reviser_component}.

From our observations, the CCR scores of the variants show a decline compared to the default input of the TAR reviser on both datasets. Interestingly, the omission of the original constraints has a more pronounced impact on CCR. This could be attributed to the negative modification of completed constraints made by the reviser when it is inaccessible to the given constraints. Additionally, when both elements are excluded, there's no noticeable difference in the CCR before and after the revision process.

Furthermore, as shown by `+ Detected by LLM' in Table \ref{tab:reviser_component}, using LLMs to detect uncompleted constraints and then feeding them back to the reviser may degrade translation performance. Further analysis of the results reveals challenges faced by LLMs inaccurately identifying unsatisfied constraints; they often mistakenly believe certain constraints have been met. Such inaccurate feedback not only fails to enhance the quality of the translation but might even deteriorate it. These insights emphasize the criticality of supplying the reviser with exact and thorough constraint information.

\section{Analysis}

\subsection{TAR Augments NMT Translators} \label{TAR_nmt}

\input{tables/nmt_base_tar}

In our study, we initially depended on constraint-aware translators to produce preliminary translation results. However, in real-world scenarios, industry practitioners often possess powerful domain-agnostic NMT models. These models, due to their lack of training with specific constraints, frequently fall short in constrained translation tasks. In this section, we integrate TAR into these general-purpose NMT models. By iteratively optimizing the NMT translation results through TAR, we can significantly enhance the CCR of the translation while ensuring its quality.

Specifically, we first the WMT21 champion model~\cite{tran2021facebook} to obtain a preliminary translation result. Since this model is not specifically trained for constraints, the initial translation often exhibits a suboptimal CCR. Building on this, we apply TAR to revise this outcome, iteratively optimizing to form the final translation result.


\begin{wrapfigure}{r}{0.45\textwidth}
\centering
\vspace{-2em}
\input{figs/different_llm1}
\caption{TAR results on WMT21 TT using \texttt{Qwen}, \texttt{ChatGPT}, \texttt{GPT-3 (text-davinci-003)} and \texttt{GPT-4}. Here ``w/o TAR'' represents the use of the conventional translation prompt. ``TAR w/o revision'' indicates the use of a prompt with constraints, but without reviser. Meanwhile, ``TAR'' denotes the full method that includes revisions.}
\label{fig:different_llms}
\vspace{-2em}
\end{wrapfigure}
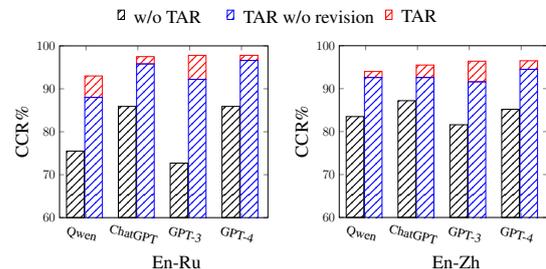

The experimental results on WMT21 TT and ETC datasets are presented in Table \ref{tab:nmt_base_tar}, we can see that TAR bolsters both BLEU and CCR scores of NMT models. On average, we observed an uplift of 0.7 and 0.8 in BLEU scores, coupled with impressive gains of 10.7\% and 14.3\% in CCR across the two datasets and language pairs. Although there remains a gap in constraint-based translation accuracy compared to the standard TAR, it generally exhibits superior translation quality. This insight demonstrates that when equipped with TAR, even domain-agnostic NMT models can adeptly tackle constrained translation. This eliminates the need for forced decoding algorithms or additional training, greatly enhancing their usability in constrained translation applications.

\subsection{Scaling TAR to More LLMs} \label{TAR_LLMs}

To evaluate the scalability and robustness of TAR across different models, we applied it to a variety of LLMs, including commercial models like \texttt{GPT-3} and \texttt{GPT-4}, as well as the open-source \texttt{Qwen}\footnote{Qwen-14B-Chat}~\cite{bai2023qwen}. We maintained consistency in all other settings.  Experiments were conducted in the "En-Zh" and "En-Ru" language directions for terminology translation. As shown in Figure \ref{fig:different_llms}, TAR consistently improves the performance of various LLMs in constrained translation tasks. To be specific, the average CCR for \texttt{GPT-3}, \texttt{GPT-4}, and \texttt{Qwen} increases by 20\%, 11.7\%, and 14.4\%, respectively. Comparing the results of \texttt{ChatGPT} with those of \texttt{GPT-3} and \texttt{GPT-4}, it's evident that TAR enables more powerful models to fully harness their capabilities in constrained translation. Intriguingly, although the CCR score of \texttt{GPT-3} in the initial translation substantially trails that of \texttt{ChatGPT}, it surpasses \texttt{ChatGPT} post-revision. While the performance of \texttt{Qwen} lags slightly behind \texttt{ChatGPT}, the improvement brought by TAR is still notable.

\subsection{Revision Iterative Round and Prompt Ensemble}
The reviser, in its function, takes the translation result and uncompleted constraints as input. Naturally, one might consider iteratively revising the output multiple times. The question arises: how many iterations strike the optimal balance? Here, we employed constraint-aware LLMs and NMT on the ``En-Zh'' and ``En-Ru'' directions of the terminology translation dataset. We assessed performance across different iterative rounds of the revision module, consistently using the same prompts for each iterative phase. 

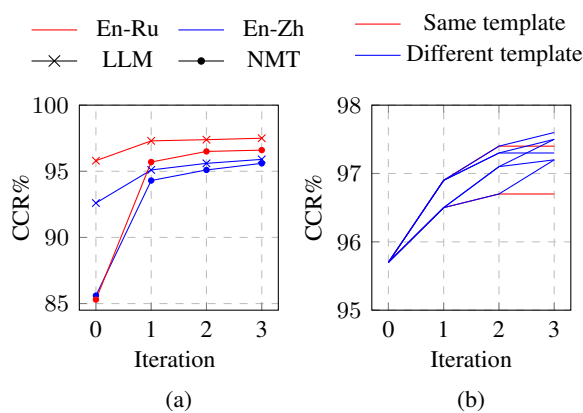
\begin{wrapfigure}{r}{0.48\textwidth}
\input{figs/round_and_template}
\caption{\textbf{(a)} Improvements in CCR with each iteration. \textbf{(b)} Red denotes consistent template use across three iterations, while blue indicates alternating templates.}
\vspace{-1.1em}
\label{fig:round_and_template}
\end{wrapfigure}


As presented in Figure \ref{fig:round_and_template_subplot:left}, there is a significant leap in performance predominantly during the initial revision. Although performance does enhance with increasing iterations, the rate of improvement starts to taper off, indicating diminishing returns.

Furthermore, we also investigated whether the LLM reviser can benefit from varying prompts in multiple iterations. To experiment this, after designing several revision templates, we randomly select one for each iterative round. Thus, a complete multi-round revision process can utilize various templates. This design is in part similar to prompt ensemble methods \cite{DBLP:journals/corr/abs-2308-12033,DBLP:journals/corr/abs-2304-05970}, combining benefits of various prompts, akin to ensemble learning \cite{DBLP:journals/fcsc/DongYCSM20}. Experimental results are shown in Figure \ref{fig:round_and_template_subplot:right}. Compared to applying a single template in revision iterations, utilizing diverse templates achieves superior performance.

\subsection{More Analysis}
Due to space limitations, we provide a more detailed analysis of our method in the appendix, including the additional costs incurred by TAR, the impact on performance as the number of constraints increases, reasons for potential declines in BLEU scores during the revision phase for certain datasets, and the performance of traditional constrained translation data augmentation methods on LLMs.


\section{Related Work}

\subsection{Constrained Translation}
Machine translation has made considerable progress in incorporating pre-specified constraint, which can be categorized into hard constrained translation and soft constrained translation. 

\textbf{Hard constrained translation:} 
This line of research expanded the original search space via decoding algorithm modification to strictly incorporate constraints~\cite{hokamp2017lexically,post2018fast,hu2019improved}. However, while these methods achieve a high constraint-based translation accuracy, they tend to be computationally expensive and can sometimes compromise translation quality~\cite{hasler2018neural,zhang2021neural}.

\textbf{Soft constrained translation:} 
Here, research primarily centers on data augmentation strategies to train NMT models to integrate constraints. Several techniques have been proposed, including: replacing source language constraints with special token~\cite{crego2016systran,wang2017sogou,zhang2023understanding}; substituting source language constraints with target language constraints~\cite{song2019code,dinu2019training}; using inline annotations to individually mark source and target language constraints~\cite{ailem2021encouraging,chen2021lexical}; and ~\cite{wang2022template} employing a template to transform the constrained translation into constraint reordering. There are also studies that modified the model architecture to better integrate vectorized constraints representation~\cite{li2020neural,wang2022integrating}, alignment information~\cite{chen2021lexically}. These approaches heavily rely on data quality or necessitate structural modifications, limiting their practicality. Moreover, they often falter when addressing diverse real-world requirements. In contrast, TAR doesn't require training on constraint-specific data and is adept at handling varied constraint scenarios.

\subsection{Automatic Post-editing}
Several studies have aimed to develop neural-based models for automatic post-editing (APE) in translation~\cite{vu2018automatic,correia2019simple,shterionov2020roadmap}. Chatterjee\shortcite{chatterjee2019automatic} investigated the application of deep learning techniques for APE and introduced novel architectures to improve the quality of post-edited translations. G{\'o}is\shortcite{gois2020learning} examined the application of automated ordering methods to improve translations. Voita\shortcite{voita2019context} introduced a context-aware approach to APE, integrating source context information into the neural framework to produce improved post-edits. Chollampatt\shortcite{chollampatt2020can} investigated the application of APE in enhancing the translation performance of NMT models. These methods primarily focus on enhancing the overall translation quality. However, it's crucial to understand that not all words within a sentence carry equal importance. The precise translation of terminologies and entities significantly impacts user experience. Our proposed TAR specifically addresses the challenge of ensuring more accurate translations for these constraints.

\section{Conclusion}

In this work, we introduce the TAR prompting method, adeptly leverages LLMs for constrained translation. Our approach involves a two-step process: first using LLMs for constrained translation, and subsequently deploying them to revise translations with uncompleted constraints. Our approach mainly improves the constraint accuracy while maintaining translation quality by overcoming the `memo trap' from the LLMs during translation using dedicated revision prompts in an iterative manner. 

We further show that TAR can be applied to LLM based translation systems as well as traditional NMT systems, in both cases resulting in better constraint accuracy while maintaining translation quality and the technology is not limited to particular LLMs. More generally, our study sheds light on the importance of accurate feedback in general for LLM revision to work effectively. 


\section{Limitations}
While we have demonstrated TAR's efficacy across four constrained translation datasets, real-world applications are considerably more varied, our prompts might not always yield optimal outcomes. In fact, the essence of TAR lies in its revision mechanism. However, as emphasized in Section \ref{sec:reviser_input}, detecting constraint adherence using LLMs poses challenges. Rule-based methods, though effective in offering accurate feedback to the reviser, can falter in broader constraint scenarios, such as controlled text generation demanding specific stylistic alignment. In such contexts, devising a method to secure accurate and efficient feedback to guide model revisions remains a research imperative. We believe that overcoming these challenges will solidify TAR's standing as a universally effective framework across diverse constraint scenarios.

\section*{Acknowledgements}

This work was supported in part by the National Science Foundation of China (No.62276056), the Natural Science Foundation of Liaoning Province of China (2022-KF-16-01), the Fundamental Research Funds for the Central Universities (Nos. N2216016 and N2316002), the Yunnan Fundamental Research Projects (No. 202401BC070021), and the Program of Introducing Talents of Discipline to Universities, Plan 111 (No.B16009).


\bibliographystyle{ccl}
\bibliography{ccl2024}

\appendix
\section{Cost of Iterations} 
\label{appendix_a}

After undergoing revisions, our TAR significantly enhances the performance of translation systems in constrained translation scenarios. However, multiple rounds of iteration introduce additional computational and financial costs. To quantitatively assess these extra expenditures, we conducted evaluations on the WMT21 terminology translation dataset. Results of time cost are shown in Table \ref{tab:aaa}, and the monetary costs are shown in Table \ref{tab:bbb}. We repeated these tests 10 times for each stage and reported the average scores. 

\input{tables/f3_and_f4}
Our results indicate a trend of diminishing returns beyond the third iteration, both in terms of performance and cost-efficiency. Specifically, the time cost stabilizes at 10 seconds per data point and the monetary expense at approximately \$4.5 per 1,000 data points by this iteration. Considering that TAR can significantly enhance the CCR in constrained translations , we believe that the cost is completely within an acceptable range.

\section{Impact of the number of constraints}
\label{appendix_b}

\begin{wrapfigure}{r}{0.48\textwidth}
\vspace{-2em}
\input{figs/number_of_contraints}
\caption{(a) The impact of increasing the number of constraints on BLEU. (b) The effect of increasing the number of constraints on CCR.}
\vspace{-1em}
\label{number_of_contraints}
\end{wrapfigure}
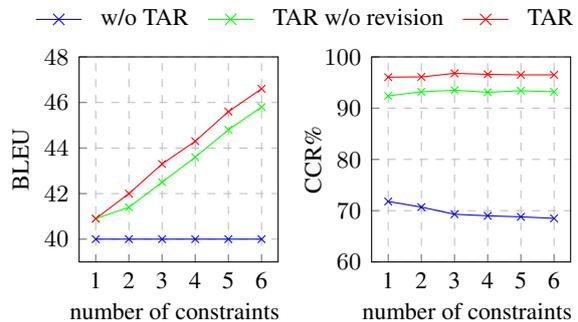
In conventional constrained translation methods, it has been observed that as the number of constraints within a single sentence increases, the CCR shows a decreasing trend. To investigate whether our method encounters the same challenge, we conducted experiments on the RTT \cite{zhang2023understanding} dataset. This dataset is composed of 500 samples meticulously selected by linguistic experts from the WMT 13-18 English-German translation test sets, with each sample being accompanied by at least six constraints. These constraints were chosen from a carefully curated set of noun phrases (e.g., names of organizations, individuals, movies, and brands) and common expressions.

To simulate different numbers of constraints, we adopted a method similar to that described in \cite{zhang2023understanding}, assuming each sentence in the test set corresponds to $N$ constraints, we randomly selected between $1$ to $N$ constraints for testing. Consequently, we constructed $k$ test subsets, with the number of constraints ranging from $1$ to $k$, Figure \ref{number_of_contraints} presents the results for two metrics ($BLEU, CCR$) as the number of constraints ($k=6$) varies. From the results, we can observe:


(1) As the number of constraints increases, the BLEU score of a standard LLM does not change, but the BLEU of both our proposed constraint-aware translator and TAR show a certain degree of improvement. When the number of constraints is 6, the BLEU of both methods can increase by about 5.5 points. This phenomenon is understandable because providing more constraints also means that more key parts of the translation sentence are already perceived by the model on how they should be translated.

(2) Similar to traditional constrained translation methods, using LLM as a translator alone, its CCR decreases as the number of constraints increases. However, the CCR of our proposed method does not change significantly, which also indicates that TAR can handle a greater number of constrained translation scenarios in real-world contexts.


\input{tables/new_table_1}

\input{tables/new_table_2}
\section{Analysis of Causes for BLEU Score Decline During the Revision Stage}
\label{appendix_c}

We observed that, among all twelve language directions, three exhibited a decline in BLEU scores after the revision stage. Theoretically, completing more constraints correctly during revision should lead to an increase in BLEU, a string-matching-based metric. However, in reality, to fulfill specific constraints, the model may employ different words or rephrase entire sentences to ensure semantic coherence, which can result in a decrease in BLEU scores. To investigate whether the revision stage could potentially degrade translation quality, we measured the changes in COMET scores before and after revision for these three language directions. The results, as shown in Table \ref{tab:new_1}, indicate that although BLEU scores declined, the COMET scores remained stable, suggesting that TAR does not compromise overall translation quality. We provide a case study in Table \ref{tab:new_2} to further illustrate this phenomenon.

%
\section{Applying mainstream constraint translation methods to LLMs}
\label{appendix_d}

In the NMT era, many studies have explored training NMT models with data augmentation to develop constrained translation capabilities. Representative approaches include Code-switching, Append, etc. To investigate the effectiveness of these methods on LLMs, we performed experiments using the IATE and Wiktionary datasets. The results are presented in Table \ref{tab:new_3}. Although code-switching prompts are somewhat effective, they typically decrease translation quality and worsen CCR metrics compared to natural language prompts. We speculate that this is because LLMs have not encountered similar prompt formats during training, leading to alignment issues during task execution. Adopting techniques such as few-shot learning and fine-tuning may mitigate issues of misalignment between prompts and model training data. However, our proposed TAR method uses natural language prompts, which are more suitable for the processing style of large language models, thus enabling better understanding and execution of translation tasks.

\input{tables/new_table_3}

\end{document}

%% file: tables/prompt.tex
\begin{tikzpicture}
\useasboundingbox (-10em,-13em) rectangle (10em,3em);
\tikzstyle{every node}=[font=\small]
\tikzstyle{modelnode} = [rectangle,draw=none,rounded corners=2pt,inner sep=2pt,minimum height=4.5em,minimum width=2em,font=\small,rotate=90]
\node [draw=black!40,inner sep=0.4em,rounded corners=4pt,dashed, thick,minimum width=42em,minimum height=6em] (it) at (0,0){};
\node [draw=black!40,inner sep=0.4em,rounded corners=4pt,dashed, thick,minimum width=42em,minimum height=9.2em] (it_1) at ([xshift=0em,yshift=-4.8em]it.south){};

\node [modelnode,anchor=north,align=center] (a) at ([xshift=-22em,yshift=-3em]it.north){\textbf{Translate}};

\node (b)[anchor=north,align=center,rounded corners=2pt,inner sep=2pt] at ([xshift=12em,yshift=2.3em]a.north)
{Translate the sentence from \textit{[src-lang]} to \textit{[tgt-lang]}, };

\node (b_0)[anchor=north,align=left,fill=prompt_blue,rounded corners=2pt,inner sep=2pt] at ([xshift=19.5em,yshift=0em]b.north){ensuring the provided constraints are reflected in the translation.};

\node (b_1) [anchor=west,align=left,rounded corners=2pt,inner sep=2pt] at ([xshift=0em,yshift=-1em]b.west)
{The constraints are given in no specific order. Only provide the translation result.};

\node (b_2) [anchor=west,align=left,rounded corners=2pt,inner sep=2pt] at ([xshift=0em,yshift=-0.8em]b_1.west)
{Sentence: $X$};

\node (b_3) [anchor=west,align=center,fill=prompt_blue,rounded corners=2pt,inner sep=2pt] at ([xshift=0em,yshift=-1em]b_2.west){Constraints:$\langle S,T \rangle$};
\node (b_4) [anchor=west,align=center,rounded corners=2pt,inner sep=2pt] at ([xshift=0em,yshift=-1em]b_3.west)
{Output:};




\node (c) [anchor=west,modelnode,align=center] at ([xshift=0em,yshift=-7.5em]a.west){\textbf{Revise}};

\node (d)[anchor=west,align=center,rounded corners=2pt,inner sep=2pt] at ([xshift=0em,yshift=-6.3em]b.west)
{Given a sentence in \textit{[src-lang]}, its constraints, and its current translation in \textit{[tgt-lang]}:};

\node (d_1)[anchor=west,align=center,rounded corners=2pt,inner sep=2pt] at ([xshift=0em,yshift=-1em]d.west)
{Original \textit{[src-lang]} sentence: $X$};

\node (d_2)[anchor=west,align=center,rounded corners=2pt,inner sep=2pt] at ([xshift=0em,yshift=-1em]d_1.west)
{Constraints: $\langle S,T \rangle$};

\node (d_3)[anchor=west,align=center,rounded corners=2pt,inner sep=2pt] at ([xshift=0em,yshift=-0.8em]d_2.west)
{Current translation: $Y^{flawed}$};

\node (d_4)[anchor=west,align=center,rounded corners=2pt,fill=orange!30,inner sep=2pt] at ([xshift=0em,yshift=-1.2em]d_3.west){Please provide a revised translation based on the following error message, ensuring that all the constraints are accur-};

\node (d_4_1)[anchor=west,align=center,rounded corners=2pt,fill=orange!30,inner sep=2pt] at ([xshift=0em,yshift=-1em]d_4.west){ately reflected in the translation:};

\node (d_5)[anchor=west,align=center,rounded corners=2pt,fill=orange!30,inner sep=2pt] at ([xshift=0em,yshift=-1em]d_4_1.west)
{Uncompleted constraints: ${\langle S,T \rangle}^{un}$};

\node (d_6)[anchor=west,align=center,rounded corners=2pt,inner sep=2pt] at ([xshift=0em,yshift=-1em]d_5.west)
{Revised translation result:};

\end{tikzpicture} 

%% file: tables/data_description.tex
\begin{wraptable}{r}{0.45\textwidth}

\small
    \centering
\vspace{-1.5em} 
 \resizebox{\linewidth}{!}{
    \begin{tabular}{c|c|c|c|c}
        \toprule
        \multirow{2}{*}{\textbf{Corpus}} & \textbf{Language} & \multirow{2}{*}{\textbf{\#Sent.}} & \textbf{\#Lines with} & \multirow{2}{*}{\textbf{\#Const.}} \\
        & \textbf{direction} & & \textbf{const.} & \\
        \midrule
        \multicolumn{5}{c}{\textit{Lexical constraint}} \\
        \midrule
        \textbf{IATE} & En-De & 2000 & 414 & 452 \\
        \textbf{Wiktionary} & En-De & 2000 & 727 & 884 \\
        \midrule
        \multirow{2}{*}{\textbf{WMT21 TT}}  & En-Ru & 2100 & 1307 & 2524 \\
                               & En-Zh & 2100 & 1191 & 2229 \\
        \midrule
        \multirow{4}{*}{\textbf{ETC}} & En-Zh & 19144 & 12040 & 35253 \\
         & En-Ru & 12985 & 3917 & 10308 \\
         & Zh-En & 19144 & 12040 & 35253 \\
         & Ru-En & 12985 & 3917 & 10308 \\
        \midrule
        \multicolumn{5}{c}{\textit{Structural constraint}} \\
        \midrule
        \multirow{4}{*}{\textbf{LXM}}  & En-Zh & 2000 & 518 & 884 \\
         & En-De & 2000 & 520 & 942 \\
         & En-Ru & 2000 & 554 & 993 \\
         & En-Fr & 2000 & 575 & 1051 \\
        \bottomrule
    \end{tabular} 
    }

    \captionof{table}{Statistics of the datasets we used for four different task: the total number of sentences in datasets (\textit{\#Sent.}), the number of lines with constraints (\textit{\#Lines with const.}), the number of constraints (\textit{\#const.}).}
    \vspace{-1em} 
    \label{tab:data_description}
    


\end{wraptable}

%% file: tables/f1_and_f2.tex
\begin{table}[t!]
\small
\centering
\begin{minipage}{1.0\textwidth}
\begin{minipage}{0.48\textwidth}
    \resizebox{1.0\linewidth}{!}{
    \begin{tabular}{l|ll|ll}
        \toprule
        \textbf{\large Method} & \textbf{\large BLEU} & \textbf{\large CCR\%} & \textbf{\large BLEU} & \textbf{\large CCR\%} \\
        \midrule
        \textit{Dataset} & \multicolumn{2}{c|}{\textit{IATE}} & \multicolumn{2}{c}{\textit{Wiktionary}} \\
        \midrule
        Transformer        & 25.8 & 76.3 & 26.0 & 76.9 \\
        Const. Dec.  & 25.3 & 82.0 & 25.8 & \textbf{99.5} \\
        Code-switching     & 26.0 & 94.5 & 26.3 & 93.4  \\
        Append     & 26.0 & 92.8 & 26.9 & 90.7  \\
        RTT & 27.2 & \textbf{99.6} & 27.8 & 98.3  \\
        LLM Trans.     & \textbf{32.0} & 85.4 & \textit{32.0} & 88.7  \\
        \midrule
        LLM Const. Trans. & \textbf{32.0} & 96.2 & \textbf{32.1} & 97.2  \\
        \rowcolor{black!20}
        \quad +Revision & \textbf{32.0\textsubscript{(+0.0)}} & \textit{98.9\textsubscript{(+2.7)}}  & \textit{32.0\textsubscript{(-0.1)}}  & \textit{98.9\textsubscript{(+1.7)}}  \\
        \bottomrule
    \end{tabular} 
    }
    \caption{Results of the general lexically constrained translation task. The \textbf{highest} scores among the various systems are highlighted in bold, while the \textit{second-best} scores are emphasized in italics for clarity.}
    \label{tab:dinu_res}
    \vspace{-1.1em}
\end{minipage}
\hfill
\begin{minipage}{0.48\textwidth}
    \resizebox{1.0\linewidth}{!}{
    \begin{tabular}{l|ll|ll}
        \toprule
        \textbf{\large Method} & \textbf{\large BLEU} & \textbf{\large CCR\%} & \textbf{\large BLEU} & \textbf{\large CCR\%} \\
        \midrule
        \textit{Direction} & \multicolumn{2}{c|}{\textit{English-Chinese}} & \multicolumn{2}{c}{\textit{English-Russian}} \\
        \midrule
        HW-TSC        & \textbf{40.7} & 88.6 & - & - \\
        TermMind-sys2  & \textit{40.5} & 85.6 & - & - \\
        ProMT.soft     & - & - & \textbf{31.1} & 90.9  \\
        TildeMT        & - & - & 28.2 & 86.3  \\
        Lingua Custodia & 29.6 & 82.8 & 28.8 & 85.4  \\
        LLM Trans.     & 36.3 & 87.2 & 29.7 & 85.9  \\
        \midrule
        LLM Const. Trans. & 36.4 & \textit{92.6} & 30.1 & \textit{95.8}  \\
        \rowcolor{black!20}
        \quad +Revision & 35.9\textsubscript{(-0.5)} & \textbf{95.9\textsubscript{(+3.3)}}  & \textit{30.3\textsubscript{(+0.2)}}  & \textbf{97.5\textsubscript{(+1.7)}}  \\
        \bottomrule
    \end{tabular} 
    }
    \caption{Results of the terminology translation task for both English-Chinese and English-Russian. }
    \label{tab:term_res}
    \vspace{1.1em}
\end{minipage}
\end{minipage}
\vspace{-1em}
\end{table}

%% file: tables/entity_res.tex
\begin{table*}[t!]
\small
    \centering

\resizebox{0.8\linewidth}{!}{
    \begin{tabular}{l|lll|lll}
        \toprule
        \textbf{\large Method} & \textbf{\large BLEU} & \textbf{\large COMET} & \textbf{\large CCR\%} & \textbf{\large BLEU} & \textbf{\large COMET} & \textbf{\large CCR\%} \\
        \midrule
        \textit{Direction} & \multicolumn{3}{c|}{\textit{English-Chinese}} & \multicolumn{3}{c}{\textit{Chinese-English}} \\
        \midrule
        Transformer  & 26.3 & 34.8 & 57.3 & 27.5 & 41.5 & 59.0 \\
        Code-switching & 25.9 & 41.4 & 70.5 & 27.2 & 45.0 & 71.1 \\
        Placeholder  & 26.4 & 42.9 & 71.4 & 27.5 & 47.2 & 72.1 \\
        Extract \& Attend & 26.8 & 48.6 & 72.3 & \textit{28.0} & 50.1 & 72.5  \\
        LLM Trans.   & 39.5 & \textbf{87.6} & 77.2 & 27.3 & \textbf{83.8} & 85.3  \\
        \midrule
        LLM Const. Trans. & \textbf{40.0} & \textit{87.4} & \textit{96.6} & \textbf{28.9} & \textit{83.5} & \textit{94.2}  \\
        \rowcolor{black!20}
        \quad +Revision & \textbf{40.0\textsubscript{(+0.0)}} & \textit{87.4\textsubscript{(+0.0)}} & \textbf{97.6\textsubscript{(+1.0)}}  & \textbf{28.9\textsubscript{(+0.0)}}  & \textit{83.5\textsubscript{(+0.0)}} & \textbf{97.3\textsubscript{(+3.1)}} \\
        

        \midrule
        
        \textit{Direction} & \multicolumn{3}{c|}{\textit{English-Russian}} & \multicolumn{3}{c}{\textit{Russian-English}} \\
        \midrule
        Transformer  & 31.8 & 52.2 & 40.0 & 34.6 & 54.0 & 48.7 \\
        Code-switching & 30.5 & 55.2 & 50.4 & 32.0 & 56.7 & 50.2 \\
        Placeholder  & 31.9 & 57.6 & 50.3 & 34.7 & 59.1 & 50.7 \\
        Extract \& Attend & \textbf{32.7} & 62.2 & 57.3 & 35.4 & 63.5 & 58.4  \\
        LLM Trans.   & 31.8 & \textbf{89.9} & 64.5 & \textit{36.0} & \textbf{85.8} & 76.8  \\
        \midrule
        LLM Const. Trans. & \textit{32.6} & \textit{89.8} & \textit{88.8} & \textbf{36.8} & \textbf{85.8} & \textit{96.7}  \\
        \rowcolor{black!20}
        \quad +Revision & 32.5\textsubscript{(-0.1)} & \textit{89.8\textsubscript{(+0.0)}} & \textbf{89.8\textsubscript{(+1.0)}} & \textbf{36.8\textsubscript{(+0.0)}} & \textbf{85.8\textsubscript{(+0.0)}} & \textbf{97.5\textsubscript{(+0.8)}} \\
        

        \bottomrule
    \end{tabular} 
   }

    \caption{Results of the entity translation task for English-Chinese, English-Russian, Chinese-English and Russian-English. }

    \label{tab:entity_res}
    

\end{table*}

%% file: tables/struct_res.tex
\begin{table*}[t!]
\small
    \centering

\resizebox{0.8\linewidth}{!}{
    \begin{tabular}{l|lll|lll}
        \toprule
        \textbf{\large Method} & \textbf{\large BLEU} & \textbf{\large SAR\%} & \textbf{\large SMR\%} & \textbf{\large BLEU} & \textbf{\large SAR\%} & \textbf{\large SMR\%} \\
        \midrule
        \textit{Direction} & \multicolumn{3}{c|}{\textit{English-Chinese}} & \multicolumn{3}{c}{\textit{English-German}} \\
        \midrule
        Transformer   & \textit{61.2} & 99.85 & 99.25 & \textit{52.7} & 99.80 & 99.20 \\
        Split-Inject  & 57.0 & \textbf{100.00} & 99.30 & 50.7 & \textbf{100.00} & \textbf{99.80} \\
        Template   & \textbf{61.5} & \textbf{100.00} & \textbf{99.80} & \textbf{53.6} & \textbf{100.00} & \textbf{99.80} \\
        LLM Trans.   & 55.1 & \textit{99.95} & 98.95 & 49.2 & \textit{99.95} & 99.25  \\
        \midrule
        LLM Const. Trans. & 56.4 & \textbf{100.00} & 99.50 & 49.2 & \textit{99.95} & 99.25  \\
        \rowcolor{black!20}
        \quad +Revision & 56.5\textsubscript{(+0.1)} & \textbf{100.00\textsubscript{(+0.00)}} & \textit{99.75}\textsubscript{(+0.25)} & 49.2\textsubscript{(+0.0)} & \textbf{100.00\textsubscript{(+0.05)}} & 99.30\textsubscript{(+0.05)} \\
        

        \midrule
        
        \textit{Direction} & \multicolumn{3}{c|}{\textit{English-French}} & \multicolumn{3}{c}{\textit{English-Russian}} \\
        \midrule
        Transformer   & \textit{65.3} & 99.55 & 99.30 & \textit{44.9
        } & 99.45 & 98.90 \\
        Split-Inject  & 66.1 & \textbf{100.00} & \textbf{100.00} & 43.1 & \textbf{100.00} & 99.85 \\
        Template   & \textbf{67.3} & \textbf{100.00} & \textbf{100.00} & \textbf{45.8} & \textbf{100.00} & \textbf{99.80} \\
        LLM Trans.   & 58.1 & \textit{99.90} & 99.30 & 34.4 & \textit{99.90} & 99.35 \\
        \midrule
        LLM Const. Trans. & 59.3 & \textbf{100.00} & \textit{99.95} & 36.0 & \textbf{100.00} & 99.60 \\
        \rowcolor{black!20}
        \quad +Revision & 59.3\textsubscript{(+0.0)}& \textbf{100.00\textsubscript{(+0.00)}} & \textit{99.95}\textsubscript{(+0.00)} & 36.0\textsubscript{(+0.0)}& \textbf{100.00\textsubscript{(+0.00)}} & \textit{99.75}\textsubscript{(+0.15)} \\
        

        \bottomrule
    \end{tabular} 
   }

    \caption{Results of the structured document translation for English-Chinese, English-German, English-Chinese and English-Russian.}

    \label{tab:structure_res}
    
\vspace{-1.6em}

\end{table*}

%% file: tables/case_study.tex
\begin{wraptable}{r}{0.4\textwidth}
  \centering
  \small
  \scalebox{0.7}{
  \vspace{-1em}
  \hspace{-1em}
  \begin{tabular}{l l}
  \toprule
  \bf Constraints & {\small $\langle${\color{case_red}WHO,\color{case_blue}\begin{CJK*}{UTF8}{gbsn} 世卫组织 \end{CJK*}}$\rangle$}; {\small $\langle${\color{case_red}COVID-19,\color{case_blue}\begin{CJK*}{UTF8}{gbsn} 新型冠状病毒 \end{CJK*}}$\rangle$}    \\
  
  \cmidrule(lr){1-1} \cmidrule(lr){2-2}
  
   \multirow{2}{*}{\bf Source} &  On 11 March 2020, {\normalsize \color{case_red}WHO }characterized \\ & {\normalsize \color{case_red} COVID-19} as a pandemic.  \\
  
  \cmidrule(lr){1-1} \cmidrule(lr){2-2}
  
  \multirow{2}{*}{\bf Reference}  & 
  \begin{CJK*}{UTF8}{gbsn} 2020年3月11日， \end{CJK*}{\normalsize \color{case_blue} \begin{CJK*}{UTF8}{gbsn} 世卫组织 \end{CJK*}}\begin{CJK*}{UTF8}{gbsn} 将 \end{CJK*}{\normalsize \color{case_blue}\begin{CJK*}{UTF8}{gbsn} 新型冠状 \end{CJK*}}\\ &{\normalsize \color{case_blue}\begin{CJK*}{UTF8}{gbsn} 病毒 \end{CJK*}}\begin{CJK*}{UTF8}{gbsn} 列为 \end{CJK*}\begin{CJK*}{UTF8}{gbsn} 大流行病。 \end{CJK*} \\
  
  
  
  
  \cmidrule(lr){1-1} \cmidrule(lr){2-2}

  
   
  \multirow{2}{*}{\bf Const. Trans.} & \begin{CJK*}{UTF8}{gbsn} 2020年3月11日， \end{CJK*}{\normalsize \color{case_blue}\begin{CJK*}{UTF8}{gbsn} 世卫组织 \end{CJK*}}\begin{CJK*}{UTF8}{gbsn} 将新冠确定为 \end{CJK*}\\ & \begin{CJK*}{UTF8}{gbsn} 大流行病。 \end{CJK*}  \\
  
  \cmidrule(lr){1-1} \cmidrule(lr){2-2}
  
  \multirow{2}{*}{\bf + Revision}  & \begin{CJK*}{UTF8}{gbsn} 2020年3月11日，\end{CJK*}{\normalsize \color{case_blue}\begin{CJK*}{UTF8}{gbsn} 世卫组织 \end{CJK*}}\begin{CJK*}{UTF8}{gbsn} 将 \end{CJK*}{\normalsize \color{case_blue}\begin{CJK*}{UTF8}{gbsn} 新型冠状 \end{CJK*}}\\ & {\normalsize \color{case_blue}\begin{CJK*}{UTF8}{gbsn} 病毒 \end{CJK*}}\begin{CJK*}{UTF8}{gbsn} 定性 \end{CJK*}\begin{CJK*}{UTF8}{gbsn} 为大流行病。 \end{CJK*} \\
  
  \bottomrule
  \end{tabular}
  }
  \caption{A case study of TAR: Initially, the translator rendered ``COVID-19'' as the more prevalent ``\begin{CJK*}{UTF8}{gbsn} 新冠 \end{CJK*}'' in Chinese. With the intervention of the reviser, it was accurately translated as ``\begin{CJK*}{UTF8}{gbsn} 新型冠状病毒 \end{CJK*}'', thereby satisfying all constraints.
  }
  \label{tab:case_study}
  \vspace{-1em}
\end{wraptable}

%% file: tables/feedback_ablation.tex
\begin{wraptable}{r}{0.45\textwidth}
\centering
\vspace{-1.7em}
\resizebox{1.0\linewidth}{!}{
    \begin{tabular}{l|cc|cc}
        \toprule
        & \multicolumn{2}{c|}{\large \textbf{IATE}} & \multicolumn{2}{c}{\large \textbf{Wiktionary}} \\
        \midrule
         Setting &  BLEU & CCR\% & BLEU & CCR\% \\
        \midrule
        base                 & 32.0 & 96.2 & 32.1 & 97.2 \\
        after revise         & 32.0 & \textbf{98.9} & 32.1 & \textbf{98.9} \\
        \midrule
        - Uncompleted const.  & 32.0 & 97.4 & 32.1 & 97.9  \\
        - Original const.    & 32.0 & 97.1 & 32.0 & 97.7  \\
        - Both               & 32.0 & 96.2 & 32.1 & 97.2 \\
        + Detected by LLM & 31.9 & 95.8 & 32.0 & 97.2 \\
        \bottomrule
    \end{tabular} 
   }
    \caption{BLEU and CCR scores of ablation on supplementary feedback. ``Uncompleted constraints'' and ``Original constraints'' are parts of the input received by the reviser.}
    \label{tab:reviser_component}
\vspace{-1.1em}
\end{wraptable}

%% file: tables/nmt_base_tar.tex
\begin{wraptable}{r}{0.45\textwidth}
\vspace{-2.3em}
\small
    \centering

\resizebox{1.0\linewidth}{!}{
    \begin{tabular}{l|ll|ll}
        \toprule
        \textbf{\large } & \textbf{\large BLEU} & \textbf{\large CCR\%} & \textbf{\large BLEU} & \textbf{\large CCR\%} \\
        \midrule
        \textbf{WMT21 TT} & \multicolumn{2}{c|}{\textit{English-Chinese}} & \multicolumn{2}{c}{\textit{English-Russian}} \\
        \midrule
        TAR & 35.9 & 95.9 & 30.3 & 97.5 \\
        NMT     &   34.5    &   85.6    &   33.6  &  85.3    \\
        \rowcolor{black!20}
         + Revision &  35.2\textsubscript{(+0.7)}    &   95.6\textsubscript{(+10.0)}    &   34.0\textsubscript{(+0.4)}  &  96.6\textsubscript{(11.3)}       \\
        \midrule
        \textbf{ETC} & \multicolumn{2}{c|}{\textit{English-Chinese}} & \multicolumn{2}{c}{\textit{English-Russian}} \\
        \midrule
        TAR & 40.0 & 97.6 & 32.5 & 89.8 \\
        NMT &  41.2    &   74.5    &   40.1  &    72.3     \\
        \rowcolor{black!20}
         + Revision &  42.7\textsubscript{(+1.5)}    &   92.8\textsubscript{(+18.3)}    &  40.2\textsubscript{(+0.1)}   &    82.4\textsubscript{(+10.1)}      \\
        \bottomrule
    \end{tabular} 
   }

    \caption{Results of applying TAR to NMT on the WMT21 TT and ETC datasets for both English-Chinese and English-Russian.
    }

    \label{tab:nmt_base_tar}
    
\vspace{-1em}

\end{wraptable}

%% file: figs/different_llm1.tex
\makeatletter
\newcommand\resetstackedplotsnew{
    \makeatletter
    \pgfplots@stacked@isfirstplottrue
    \makeatother
    \addplot [forget plot,draw=none] coordinates{(Qwen, 0) (ChatGPT, 0) (GPT-3, 0) (GPT-4, 0)};
}
\begin{minipage}{0.23\textwidth}
\begin{tikzpicture} [scale=0.4]
\tikzstyle{every node}=[font=\large]
\begin{axis} [
ybar stacked,
ylabel=\LARGE CCR\%,
ymin=60,
ymax=100,
xtick = data,
xticklabel style={rotate=-14, anchor=center, xshift=0em, yshift=-1.5em},
bar width=.58cm,
enlarge x limits=0.17,
symbolic x coords={Qwen, ChatGPT, GPT-3, GPT-4},
legend style={ legend columns=-1,
    column sep = 0.6em,
    /tikz/column 2/.style={column sep=2em,},
    at={(0.9,1.1)},
    anchor=south,
    draw=none}, 
]

\addplot [bar shift=-.35cm,
    draw = black,
    semithick,
    pattern = north east lines,
    pattern color = black
] coordinates {(ChatGPT,85.9) (GPT-3,72.7) (GPT-4,85.9) (Qwen,75.5)};

\resetstackedplotsnew

\addplot [bar shift=.25cm,
    draw = blue,
    semithick,
    pattern = north east lines,
    pattern color = blue
] coordinates {(ChatGPT,95.8) (GPT-3,92.2) (GPT-4,96.6) (Qwen,88.0)};

\addplot [bar shift=.25cm,
    draw = red,
    semithick,
    pattern = north east lines,
    pattern color = red
] coordinates {(ChatGPT,1.7) (GPT-3,5.6) (GPT-4,1.2) (Qwen,5.0)};

\legend {\LARGE w/o TAR , \LARGE TAR w/o revision};
\end{axis} 

\node (lang)[anchor=north,font=\scriptsize] at ([xshift=10em,yshift=-10em]it.north){En-Ru};

\end{tikzpicture}
\end{minipage}%
\begin{minipage}{0.23\textwidth}
\begin{tikzpicture}[scale=0.4]
\tikzstyle{every node}=[font=\large]
\begin{axis} [
ybar stacked,
ylabel=\LARGE CCR\%,
ymin=60,
ymax=100,
xtick = data,
xticklabel style={rotate=-14, anchor=center, xshift=0em, yshift=-1.5em},
bar width=.58cm,
enlarge x limits=0.17,
symbolic x coords={Qwen, ChatGPT, GPT-3, GPT-4},
legend style={ legend columns=-1,
    column sep = 0.6em,
    at={(0.35,1.1)},
    anchor=south,
    draw=none}, 
]
\addplot [bar shift=-.35cm,
    draw = black,
    semithick,
    pattern = north east lines,
    pattern color = black
] coordinates {(ChatGPT,87.2) (GPT-3,81.6) (GPT-4,85.2) (Qwen,83.5)};

\resetstackedplotsnew

\addplot [bar shift=.25cm,
    draw = blue,
    semithick,
    pattern = north east lines,
    pattern color = blue
] coordinates {(ChatGPT,92.6) (GPT-3,91.6) (GPT-4,94.5) (Qwen,92.6)};

\addplot [bar shift=.25cm,
    draw = red,
    semithick,
    pattern = north east lines,
    pattern color = red
] coordinates {(ChatGPT,2.9) (GPT-3,4.8) (GPT-4,2.0) (Qwen,1.4)};

\legend {, , \LARGE TAR};
\end{axis}
\node (lang)[anchor=north,font=\scriptsize] at ([xshift=10em,yshift=-10em]it.north){En-Zh};
\end{tikzpicture}

\end{minipage}%

%% file: figs/round_and_template.tex
\newenvironment{customlegend}[1][]{%
    \begingroup
    \csname pgfplots@init@cleared@structures\endcsname
    \pgfplotsset{#1}%
}{%
    \csname pgfplots@createlegend\endcsname
    \endgroup
}%

\def\addlegendimage{\csname pgfplots@addlegendimage\endcsname}

\begin{centering}
\hspace*{-1.09em}
\begin{tikzpicture}
        \begin{customlegend}[
            legend style={
                column sep = 5pt, 
                at={(1.2,4)},
                anchor=north,
                draw=none
            },
            legend columns=2,
            legend entries={\small En-Ru, \small En-Zh, \small LLM, \small NMT}
        ]
        \addlegendimage{red,sharp plot}
        \addlegendimage{blue,sharp plot}
        
        \addlegendimage{black, mark=x, mark size=3pt}
        \addlegendimage{black, mark=*, mark size=1.2pt}
        
        \end{customlegend}
        \pgfplotsset{footnotesize,samples=2}
        
        \begin{groupplot}[
            group style = {group name=my plots, group size = 2 by 1, horizontal sep = 35pt}, ]
            \nextgroupplot[
                align = center,
                title={Iteration}, 
                title style={at={(axis description cs:0.48,-0.15)},anchor=north}, 
                legend style={ legend columns=2,
                    at={(0.5,1.5)},
                    anchor=north,
                    draw=none
                }, 
                width=4.2cm, height=4.3cm,
                ymin=84.5, ymax=100,
                symbolic x coords={0,1,2,3},
                xtick=data,
                ylabel={CCR\%},
                ylabel style={align=center},
                ytick={85, 90, 95, 100},
                x label style={at={(axis description cs:0.5,-0.15)},anchor=north},
                y label style={at={(axis description cs:0.27,0.5)},anchor=south},
                xtick pos=bottom,
                ytick pos=left,
                xmajorgrids,
                ymajorgrids,
                major grid style={dashed},
                ]
            
                \addplot[
                    color=blue,
                    mark=x,
                    mark size=2pt,
                    ]
                    coordinates {
                    (0,92.6)
                    (1,95.1)
                    (2,95.6)
                    (3,95.9)
                    };
                
                \addplot[
                    color=blue,
                    mark=*,
                    mark size=1pt,
                    ]
                    coordinates {
                    (0,85.6)
                    (1,94.3)
                    (2,95.1)
                    (3,95.6)
                    };
                \addplot[
                    color=red,
                    mark=x,
                    mark size=2pt,
                    ]
                    coordinates {
                    (0,95.8)
                    (1,97.3)
                    (2,97.39)
                    (3,97.5)
                    };
                \addplot[
                    color=red,
                    mark=*,
                    mark size=1pt,
                    ]
                    coordinates {
                    (0,85.3)
                    (1,95.7)
                    (2,96.5)
                    (3,96.6)
                    };

            \nextgroupplot[
                align = center,
                title = {Iteration},
                title style={at={(axis description cs:0.5
                ,-0.15)},anchor=north},
                width=4.2cm, height=4.3cm,
                ymin=95, ymax=98,
                symbolic x coords={0, 1, 2, 3},
                xtick=data,
                ylabel={CCR\%},
                ylabel style={align=center},
                ytick={95, 96, 97, 98},
                x label style={at={(axis description cs:0.07,-0.22)},anchor=north},
                y label style={at={(axis description cs:0.27,0.5)},anchor=south},
                legend style={ legend columns=1,
                    at={(0.5,1.5)},
                    anchor=north,
                    draw=none
                },
                ytick pos=left,
                xmajorgrids,
                ymajorgrids,
                major grid style={dashed},
                ]
                \addplot[
                    color=red,
                    mark size=2.5pt,
                    bar shift =0.1cm
                    ]
                    coordinates {
                    (0,95.7)
                    (1,96.9)
                    (2,97.4)
                    (3,97.4)
                    };
                \addplot[
                    color=blue,
                    mark size=2.5pt,
                    bar shift =0.1cm
                    ]
                    coordinates {
                    (0,95.7)
                    (1,96.9)
                    (2,97.4)
                    (3,97.6)
                    };
                
                \addplot[
                    color=blue,
                    mark size=2.5pt,
                    bar shift =0.1cm
                    ]
                    coordinates {
                    (0,95.7)
                    (1,96.9)
                    (2,97.3)
                    (3,97.5)
                    };
                \addplot[
                    color=blue,
                    mark size=2.5pt,
                    bar shift =0.1cm
                    ]
                    coordinates {
                    (0,95.7)
                    (1,96.9)
                    (2,97.3)
                    (3,97.3)
                    };

                \addplot[
                    color=red,
                    mark size=2.5pt,
                    bar shift =0.1cm
                    ]
                    coordinates {
                    (0,95.7)
                    (1,96.5)
                    (2,96.7)
                    (3,96.7)
                    };
                \addplot[
                    color=blue,
                    mark size=2.5pt,
                    bar shift =0.1cm
                    ]
                    coordinates {
                    (0,95.7)
                    (1,96.5)
                    (2,97.1)
                    (3,97.5)
                    };
                \addplot[
                    color=blue,
                    mark size=2.5pt,
                    bar shift =0.1cm
                    ]
                    coordinates {
                    (0,95.7)
                    (1,96.5)
                    (2,97.1)
                    (3,97.2)
                    };
                \addplot[
                    color=blue,
                    mark size=2.5pt,
                    bar shift =0.1cm
                    ]
                    coordinates {
                    (0,95.7)
                    (1,96.5)
                    (2,96.7)
                    (3,97.2)
                    };
                \legend{Same template, , ,Different template}
            
        \end{groupplot}
    \tikzset{SubCaption/.style={text width=2in,yshift=-1.8em, align=center,anchor=north}}
 
    \node[SubCaption] at (my plots c1r1.south) {\subcaption{\label{fig:round_and_template_subplot:left}}};
    \node[SubCaption] at (my plots c2r1.south) {\subcaption{\label{fig:round_and_template_subplot:right}}};
    
    \end{tikzpicture}
    
    \end{centering}

%% file: tables/f3_and_f4.tex
\begin{table}[t!]
\small
\centering
\begin{minipage}{1.0\textwidth}
\begin{minipage}{0.48\textwidth}
    \resizebox{1.0\linewidth}{!}{
    \begin{tabular}{l|ll|ll}
        \toprule
        \textbf{\large  } & \textbf{\large CCR\%} & \textbf{\large Cost(s) } & \textbf{\large CCR\%} & \textbf{\large Cost(s) } \\
        \midrule
        \textit{Direction} & \multicolumn{2}{c|}{\textit{English-Chinese}} & \multicolumn{2}{c}{\textit{English-Russian}} \\
        \midrule
        Iteration0  & 92.6 & 1.48 & 87.9 & 2.28 \\
        Iteration1  & 93.9 & 2.21 & 93.0 & 3.26 \\
        Iteration2  & 94.6 & 2.02 & 94.8 & 2.99 \\
        Iteration3  & 95.0 & 2.39 & 95.1 & 3.17 \\
        Iteration4  & 95.3 & 2.56 & 96.2 & 3.02 \\
        Iteration5  & 95.6 & 2.42 & 96.9 & 3.25 \\
        \bottomrule
    \end{tabular} 
    }
    \caption{The processing time for each data point at various stages using the Qwen-14b-chat model on an A100 GPU.}
    \label{tab:aaa}
\end{minipage}
\hfill
\begin{minipage}{0.48\textwidth}
    \resizebox{1.0\linewidth}{!}{
    \begin{tabular}{l|ll|ll}
        \toprule
        \textbf{\large  } & \textbf{\large CCR\%} & \textbf{\large Cost(\$) } & \textbf{\large CCR\%} & \textbf{\large Cost(\$) } \\
        \midrule
        \textit{Direction} & \multicolumn{2}{c|}{\textit{English-Chinese}} & \multicolumn{2}{c}{\textit{English-Russian}} \\
        \midrule
        Iteration0  & 92.6 & 0.62 & 95.8 & 0.75 \\
        Iteration1  & 95.1 & 1.20 & 97.0 & 1.28 \\
        Iteration2  & 95.6 & 1.26 & 97.3 & 1.08 \\
        Iteration3  & 95.9 & 1.31 & 97.5 & 1.23 \\
        Iteration4  & 95.9 & 1.34 & 97.6 & 1.23 \\
        Iteration5  & 95.9 & 1.34 & 97.6 & 1.21 \\
        \bottomrule
    \end{tabular} 
    }
    \caption{The monetary costs involved in processing every 1,000 data points using gpt3.5-turbo-0613 at different stages.}
    \label{tab:bbb}
\end{minipage}
\end{minipage}
\end{table}
\label{f3_and_f4}

%% file: figs/number_of_contraints.tex
\newenvironment{customlegend}[1][]{%
    \begingroup
    \csname pgfplots@init@cleared@structures\endcsname
    \pgfplotsset{#1}%
}{%
    \csname pgfplots@createlegend\endcsname
    \endgroup
}%

\def\addlegendimage{\csname pgfplots@addlegendimage\endcsname}

\begin{centering}
\hspace*{-1.09em}
\begin{tikzpicture}
    \begin{customlegend}[
        legend style={
            column sep = 5pt, 
            at={(3,3.5)},
            anchor=north,
            draw=none
        },
        legend columns=3,
        legend entries={\small w/o TAR, \small TAR w/o revision, \small TAR}
    ]
    \addlegendimage{blue, mark=x, mark size=3pt}
    \addlegendimage{green, mark=x, mark size=3pt}
    \addlegendimage{red, mark=x, mark size=3pt}
    
    \end{customlegend}
    \pgfplotsset{footnotesize,samples=2}
    
    \begin{groupplot}[
        group style = {group name=my plots, group size = 2 by 1, horizontal sep = 35pt}
    ]
        \nextgroupplot[
            align = center,
            title={number of constraints}, 
            title style={at={(axis description cs:0.48,-0.15)},anchor=north}, 
            legend style={ legend columns=2,
                at={(0.5,1.5)},
                anchor=north,
                draw=none
            }, 
            width=4.2cm, height=4.3cm,
            ymin=39, ymax=48,
            symbolic x coords={1,2,3,4,5,6},
            xtick=data,
            ylabel={BLEU},
            ylabel style={align=center},
            ytick={40, 42, 44, 46, 48},
            x label style={at={(axis description cs:0.5,-0.15)},anchor=north},
            y label style={at={(axis description cs:0.27,0.5)},anchor=south},
            xtick pos=bottom,
            ytick pos=left,
            xmajorgrids,
            ymajorgrids,
            major grid style={dashed},
        ]

            \addplot[
                color=blue,
                mark=x,
                mark size=2pt,
            ]
                coordinates {
                (1,40)
                (2,40)
                (3,40)
                (4,40)
                (5,40)
                (6,40)
                };
            \addplot[
                color=green,
                mark=x,
                mark size=2pt,
            ]
                coordinates {
                (1,40.9)
                (2,41.4)
                (3,42.5)
                (4,43.6)
                (5,44.8)
                (6,45.8)
                };
            \addplot[
                color=red,
                mark=x,
                mark size=2pt,
            ]
                coordinates {
                (1,40.9)
                (2,42)
                (3,43.3)
                (4,44.3)
                (5,45.6)
                (6,46.6)
                };

        \nextgroupplot[
            align = center,
            title={number of constraints}, 
            title style={at={(axis description cs:0.48,-0.15)},anchor=north}, 
            legend style={ legend columns=2,
                at={(0.5,1.5)},
                anchor=north,
                draw=none
            }, 
            width=4.2cm, height=4.3cm,
            ymin=60, ymax=100,
            symbolic x coords={1,2,3,4,5,6},
            xtick=data,
            ylabel={CCR\%},
            ylabel style={align=center},
            ytick={60, 70, 80, 90, 100},
            x label style={at={(axis description cs:0.5,-0.15)},anchor=north},
            y label style={at={(axis description cs:0.27,0.5)},anchor=south},
            xtick pos=bottom,
            ytick pos=left,
            xmajorgrids,
            ymajorgrids,
            major grid style={dashed},
        ]

            \addplot[
                color=blue,
                mark=x,
                mark size=2pt,
            ]
                coordinates {
                (1,71.8)
                (2,70.7)
                (3,69.3)
                (4,69.0)
                (5,68.8)
                (6,68.5)
                };
            \addplot[
                color=green,
                mark=x,
                mark size=2pt,
            ]
                coordinates {
                (1,92.4)
                (2,93.2)
                (3,93.5)
                (4,93.1)
                (5,93.4)
                (6,93.2)
                };
            \addplot[
                color=red,
                mark=x,
                mark size=2pt,
            ]
                coordinates {
                (1,96.05)
                (2,96.1)
                (3,96.8)
                (4,96.6)
                (5,96.5)
                (6,96.5)
                };
    \end{groupplot}
\tikzset{SubCaption/.style={text width=2in,yshift=-1.8em, align=center,anchor=north}}

\label{number_of_constraints}
\end{tikzpicture}

\end{centering}

%% file: tables/new_table_1.tex
\begin{table}[t!]
\small
\centering
    \resizebox{0.5\linewidth}{!}{
    \begin{tabular}{l|lll}
        \toprule
        \textbf{Wikitionary} & \textbf{BLEU} & \textbf{COMET} & \textbf{CCR\%} \\
        \arrayrulecolor{black!40} \cmidrule(lr){1-4} \arrayrulecolor{black}
        LLM Const. Trans. & 32.1 & 87.3 & 97.2 \\
        \quad +Revision & 32.0 & 87.3 & 98.9 \\
        \midrule
        \textbf{WMT21 TT en-zh} & \textbf{BLEU} & \textbf{COMET} & \textbf{CCR\%} \\
        \arrayrulecolor{black!40} \cmidrule(lr){1-4} \arrayrulecolor{black} 
        LLM Const. Trans. & 36.4 & 86.9 & 92.6 \\
        \quad +Revision & 35.9 & 86.9 & 95.9 \\
        \midrule
        \textbf{ETC en-ru} & \textbf{BLEU} & \textbf{COMET} & \textbf{CCR\%} \\
        \arrayrulecolor{black!40} \cmidrule(lr){1-4} \arrayrulecolor{black}
        LLM Const. Trans. & 32.6 & 89.9 & 88.8 \\
        \quad +Revision & 32.5 & 89.8 & 89.8 \\
        \bottomrule
    \end{tabular} 
    }
    \caption{Comparison of metrics before and after revision on the Wiktionary, WMT21 TT en-zh, and ETC en-ru datasets.}
    \label{tab:new_1}
    \vspace{1.1em}
\vspace{-1em}
\end{table}

%% file: tables/new_table_2.tex
\begin{table}[t!]
\small
\centering
\resizebox{1.0\linewidth}{!}{
\begin{tabular}{l|l}
    \toprule
    \textbf{Constraints} & virus spread $\rightarrow$ \begin{CJK*}{UTF8}{gbsn} 病毒传播 \end{CJK*} \& Wuhan $\rightarrow$ \begin{CJK*}{UTF8}{gbsn} 武汉 \end{CJK*} or \begin{CJK*}{UTF8}{gbsn} 武汉市 \end{CJK*} \\
    \midrule 
    Source & \makecell[l]{In early and mid-January 2020, the virus spread to other Chinese provinces, helped by the Chinese New Year migration and Wuhan being \\ a transport hub and major rail interchange.} \\
    \midrule 
    Reference & \begin{CJK*}{UTF8}{gbsn} 在2020年1月初至1月中旬，受中国春节人口大流动和武汉作为交通枢纽和主要铁路枢纽的影响，病毒传播到了中国其他省份。 \end{CJK*}  \\
    \midrule 
    Const. Trans. & \begin{CJK*}{UTF8}{gbsn}  2020年1月初和中旬，病毒通过中国春节迁徙和武汉作为交通枢纽和主要铁路换乘站的帮助，传播到其他中国省份。  \end{CJK*} \\
    \midrule 
    +Revision & \begin{CJK*}{UTF8}{gbsn} 2020年1月初和中旬，病毒传播到其他中国省份，得益于中国春节迁徙和武汉市作为交通枢纽和主要铁路换乘站的地位。 \end{CJK*} \\
    \bottomrule
\end{tabular} 
}
\caption{A case demonstrated that duringthe revsion phase,the constrain oftranslating ``virus spread" to\begin{CJK*}{UTF8}{gbsn} ``病毒传播 \end{CJK*}" was completed. Meanwhile, the model reorganized the sentence structure and adjusted the wording, resulting in a 5.25 decrease in BLEU score. However, the overall fluency did not change.}
\label{tab:new_2}
\vspace{-1.5em}
\end{table}

%% file: tables/new_table_3.tex
\begin{table}[t!]
\small
\centering
\resizebox{0.5\linewidth}{!}{
\begin{tabular}{l|ll|ll}
    \toprule
    \textbf{\large Method} & \textbf{\large BLEU} & \textbf{\large CCR\%} & \textbf{\large BLEU} & \textbf{\large CCR\%} \\
    \midrule
    \textit{Dataset} & \multicolumn{2}{c|}{\textit{IATE}} & \multicolumn{2}{c}{\textit{Wiktionary}} \\
    \midrule
    LLM Trans           & 32.0 & 85.4 & 32.0 & 88.7 \\
    LLM Code-switching  & 31.8 & 94.0 & 31.9 & 92.6 \\
    LLM Append          & 31.6 & 93.3 & 31.5 & 91.9 \\
    LLM Const. Trans.   & 32.0 & 96.2 & 32.1 & 97.2 \\
    \quad +Revision     & 32.0 & 98.9 & 32.0 & 98.9 \\
    \bottomrule
\end{tabular} 
}
\caption{Performance of traditional data augmentation methods on LLMs.}
\label{tab:new_3}
\vspace{-1.1em}

\end{table}